\documentclass[10pt,journal,compsoc]{IEEEtran}

\usepackage{xcolor,soul,framed} %,caption

\colorlet{shadecolor}{yellow}
\usepackage[pdftex]{graphicx}
\graphicspath{{../pdf/}{../jpeg/}}
\DeclareGraphicsExtensions{.pdf,.jpeg,.png}

\usepackage[cmex10]{amsmath}
\usepackage{array}
\usepackage{mdwmath}
\usepackage{mdwtab}
\usepackage{eqparbox}
\usepackage{url}
\usepackage{graphicx}
\usepackage{amsmath}
\usepackage{amsfonts}
\usepackage{algorithm}
\usepackage{algpseudocode}
\usepackage{multirow}
\usepackage{footnote}
\usepackage{bm}
\usepackage{makecell}
\usepackage{pdflscape}
\usepackage{afterpage}
\usepackage{bbding}
\usepackage{lscape}
\usepackage{booktabs}
\usepackage{tikz}

\usepackage{pgfplots}
\pgfplotsset{compat=1.12}
\usepackage{subcaption}
\pgfplotsset{
	layers/my layer set/.define layer set={
		background,
		main,
		foreground
	}{ },
	set layers=my layer set,
}

\usepackage[square,sort,comma,numbers]{natbib} % multiple citation or 
\usepackage[top=0.4in,bottom=0.4in,left=0.4in,textwidth=7.7in]{geometry}

\usepackage[normalem]{ulem}
\useunder{\uline}{\ul}{}

\usepackage[pagebackref=false,breaklinks=false,linkcolor=black,anchorcolor=black, citecolor=black,colorlinks,bookmarks=true]{hyperref}
\newcommand{\ie}{\textit{i}.\textit{e}., }

\usepackage{amsfonts}

\usepackage{pifont}

\title{SceneTracker: Long-term Scene Flow \\
	Estimation Network}

\author{\IEEEauthorblockN{
		Bo Wang\thanks{
			Bo Wang, Jian Li, Yang Yu, Zhenping Sun, and Dewen Hu are with the College of Intelligence Science and Technology, National University of Defense Technology, Changsha, China.
			Li Liu is with the College of Electronic Science and Technology, National University of Defense Technology, Changsha, China.
		}, 
		Jian Li, Yang Yu, Li Liu, Zhenping Sun, Dewen Hu
		\thanks{
			Bo Wang and Jian Li contributed equally to this work. Corresponding authors: Zhenping Sun (sunzhenping@nudt.edu.cn) and Dewen Hu (dwhu@nudt.edu.cn). 
			This work was partially supported by 
			the National Natural Science Foundation of China under Grant 61973311, 62376283, 62006239, 
			and the Science and Technology Innovation Program of Hunan Province (2024QK2006)
		}
}}

\begin{document}

\IEEEtitleabstractindextext{%
	\begin{abstract}
		Considering that scene flow estimation has the capability of the spatial domain to focus but lacks the coherence of the temporal domain, this study proposes long-term scene flow estimation (LSFE), a comprehensive task that can simultaneously capture the fine-grained and long-term 3D motion in an online manner. 
		We introduce SceneTracker, the first LSFE network that adopts an iterative approach to approximate the optimal 3D trajectory. 
		The network dynamically and simultaneously indexes and constructs appearance correlation and depth residual features. 
		Transformers are then employed to explore and utilize long-range connections within and between trajectories. 
		With detailed experiments, SceneTracker shows superior capabilities in addressing 3D spatial occlusion and depth noise interference, highly tailored to the needs of the LSFE task. 
		We build a real-world evaluation dataset, LSFDriving, for the LSFE field and use it in experiments to further demonstrate the advantage of SceneTracker in generalization abilities. 
		The code and data are available at \href{https://github.com/wwsource/SceneTracker}{https://github.com/wwsource/SceneTracker}. 
	\end{abstract}
	
	}

\maketitle
\IEEEdisplaynontitleabstractindextext
\IEEEpeerreviewmaketitle

\section{Introduction}

The precise and online capture and analysis of the fine-grained long-term 3D object motion play a pivotal role in scene comprehension. By accurately capturing the 3D movement of objects, detailed data on position, velocity, acceleration, and more can be obtained, allowing for an in-depth understanding of how objects move and interact within specific environments. This information holds critical significance across various domains, including robotics, autonomous driving, and virtual reality. 

The existing work primarily focuses on scene flow estimation (SFE), which partially addresses the issue of online 3D motion capture. 
SFE aims to estimate the pixel-wise/point-wise 3D motion between consecutive stereo, RGB-D, or point cloud data, and it can capture the fine-grained instantaneous 3D displacement. 
However, SFE is unable to robustly capture the 3D trajectory over a period of time. Due to the presence of occlusions, per-frame 3D displacements cannot form a trajectory through simple chaining. 

Considering that SFE has the capability of the spatial domain to focus but lacks the coherence of the temporal domain, this study proposes long-term scene flow estimation (LSFE), a comprehensive task that can simultaneously capture the fine-grained and long-term 3D motion in an online manner. 
LSFE estimates the 3D trajectory of the query pixel/point within a continuous stereo, RGB-D, or point cloud sequence online, and can be seen as an extension of SFE to a broader temporal domain. 

We propose SceneTracker, the first LSFE method, which takes a $T$-frame RGB-D video and camera intrinsics as input, along with the starting 3D coordinates $x, y, z$ of an arbitrary 3D target to track, and produces a $T \times 3$ matrix as output, representing the positions in the camera coordinate system of the target across the given frames. 
SceneTracker adopts an iterative approach to approximate the optimal 3D trajectory, overcoming the significant displacement challenges between frames. 
The network dynamically and simultaneously indexes and constructs appearance correlation and depth residual features, enhancing the ability to localize the target 3D position. 
Transformers are then employed to explore and utilize long-range connections within and between trajectories, further improving accuracy. 

We train and evaluate the proposed method on an augmented dataset, LSFOdyssey, which is built on the large-scale synthetic dataset, PointOdyssey \cite{pointodyssey}. 
In experiments on the LSFOdyssey dataset, SceneTracker achieves median 3D error reductions of 70.5\% and 86.6\%, respectively, compared with scene flow-based chaining and tracking any point-based depth indexing methods. 
The results demonstrate superior capabilities of SceneTracker in addressing 3D spatial occlusion and depth noise interference, highly tailored to the needs of the LSFE task. 

We build a real-world evaluation dataset, LSFDriving, for the LSFE field. The dataset is geared toward the autonomous driving realm, providing 3D trajectory annotations for points on objects with diverse motion attributes and rigidities. 
Specifically, the tracked points are sampled from static backgrounds, moving rigid vehicles, and moving non-rigid pedestrians. 
Addressing the complexity of local nonlinear motion in pedestrian joints and garments, we build a semi-automated annotation pipeline that can efficiently and accurately annotate 3D trajectories. 
In experiments on the LSFDriving dataset, SceneTracker demonstrates an advantage in generalization abilities in real-world scenes while training solely on synthetic datasets. 

The study makes the following contributions: 
\begin{itemize}
\renewcommand{\labelitemi}{$\bullet$}
\item To better capture the fine-grained and long-term 3D motion, the comprehensive task of long-term scene flow estimation is studied. 
\item The first LSFE network, SceneTracker, is presented. With detailed experiments, SceneTracker shows superior capabilities in addressing 3D spatial occlusion and depth noise interference. 
\item We build a real-world evaluation dataset, LSFDriving, for the LSFE field and use it in experiments to further demonstrate the advantage of SceneTracker in generalization abilities. 
\end{itemize}

\section{Related Works}

\subsection{Scene Flow Estimation}

Existing methods generally rely on feature similarities as their underlying principle. FlowNet3D \cite{flownet3d} operates directly on point clouds with PointNet++ \cite{pointnet++}. CamLiFlow \cite{CamLiFlow} refines the optical flow and scene flow from the similarity tensor of two frames in a coarse-to-fine way with two branches. OpticalExp \cite{opticalexp} uses optical expansion to recover the motion of a pixel in depth between frames. It can estimate the final depth value of the second frame by combining the depth of the first frame, indirectly estimating the scene flow per pixel. 
Occlusion issues severely constrain the accuracy of SFE. To address the occlusion problem, RAFT-3D \cite{raft3d} and RigidMask \cite{rigidmask} focus on introducing rigid motion priors implicitly and explicitly, respectively. 

\subsection{Tracking Any Point}

Unlike optical flow estimation methods \cite{Flownet_and_Chairs, PWCNet, hui2018liteflownet, xu2023unifying, RAFT, GMA, splatflow}, which focus on local instantaneous motion, the task of tracking any point (TAP) focuses more on the global motion trajectory in the image domain. 
PIPs \cite{pips} combines cost volumes and iterative inference with a deep temporal network to jointly reason about the locations and appearance of visual entities across multiple time-steps, achieving excellent tracking results. 
TAP-Net \cite{tapnet} independently computes cost volumes between the first frame and each intermediate frame of a video, and uses these volumes directly for regression of position and occlusion status. 
MFT \cite{mft} performs the chaining of optical flow and occlusion status by estimating the uncertainty. 
TAPIR \cite{tapir} combines the global matching strategy of TAP-Net with the refinements of PIPs. 
PIPs++ \cite{pointodyssey} introduces 1D convolution and dynamic templates, improving the tracking survival rate over longer time spans. 
Cotracker \cite{cotracker} utilizes the strong correlation between multiple tracking points for joint tracking. 
In contrast to the TAP task, LSFE overcomes inherent noise in depth data and considers the 3D structures and motion patterns of objects to accurately track trajectories in 3D space. 

\subsection{Reconstruction-based Offline Methods}

Inspired by 3D-GS \cite{3DGS}, which is notable for its pure explicit representation and differential point-based splatting method, many works have started to use 3D Gaussians to reconstruct dynamic scenes and render novel views. Dynamic 3D Gaussians \cite{dynamic3DGS} models dynamic scenes by tracking the position and variance of each 3D Gaussian at each timestamp. 4D-GS \cite{4DGS} proposes 4D Gaussian Splatting as a holistic representation for dynamic scenes rather than applying 3D-GS to each frame. After training on the complete multi-view videos, these methods calculate 3D trajectories based on Gaussian deformations. 
Unlike this offline approach, the LSFE task focuses more on challenging scenarios involving constrained observations, limited response time, and future uncertainty. 

\section{Proposed Method}

\subsection{A Review of CoTracker}

The architecture of our method is primarily inspired by the TAP method CoTracker \cite{cotracker}. The main process for CoTracker to estimate 2D trajectories is as follows:
CoTracker processes 2D video by dividing it into sliding windows, estimating 2D trajectory segments within each window, and linking them to produce final trajectories. During each sliding window processing, the network first initializes 2D trajectory segments and then iteratively updates template features and 2D trajectory segments. In each iteration, the network constructs the appearance correlation feature and employs Transformers to establish long-range connections within and between 2D trajectories. 
Given space constraints and for self-containment, a more detailed description of CoTracker is provided in the appendix. 

Our method differs from CoTracker in the following aspects: 
1) We construct the depth residual feature providing target positional information along the depth dimension; 
2) We design a flow iteration module leveraging the depth residual feature and depth motion for iterative regression of 3D trajectories; 
3) We introduce the loss function capable of supervising trajectory estimation in both the $uv$ plane and the depth dimension, in an unrolled fashion. 

\subsection{SceneTracker Architecture} \label{section: overall}

\begin{figure*}[t]
	\begin{center}
		\includegraphics[width=\linewidth]{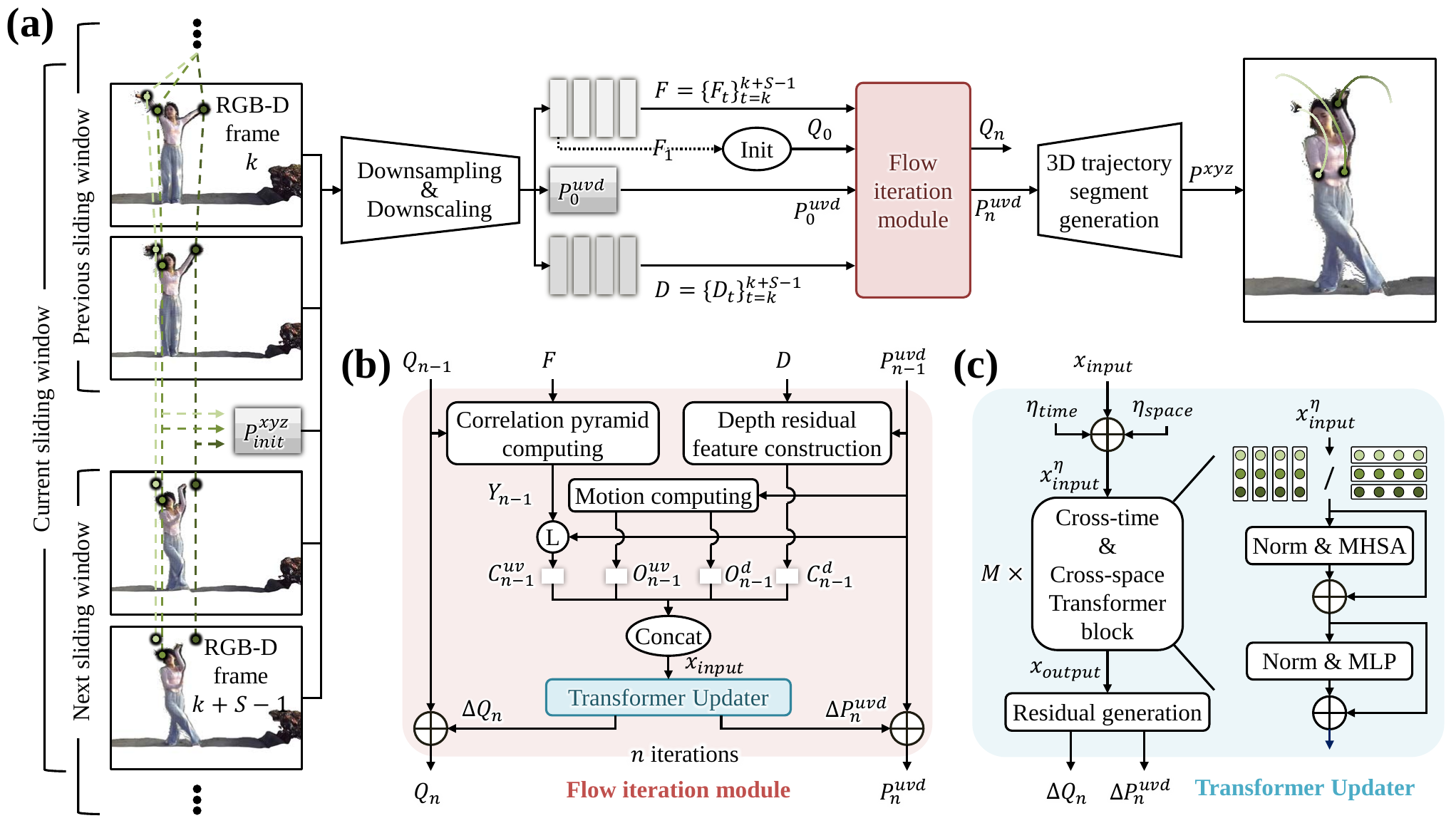}
	\end{center}
	\caption{
		\textbf{Architecture of the proposed method.} 
		\textbf{(a)} LSFE process of a current sliding window. 
		Using $S$ RGB-D frames and the initialized trajectories $P^{xyz}_{init}$ as inputs, the network estimates the 3D trajectory segments $P^{xyz}$. 
		\textbf{(b)} Flow iteration module. 
		Template feature $Q_n$ and 3D downscaled trajectories $P^{uvd}_n$ are iteratively updated. 
		\textbf{(c)} Transformer Updater network. 
		The input feature is enhanced by Transformer blocks that factorize the attention across time and space. 
	}
	\label{fig: overview}
\end{figure*}

We seek to track 3D points throughout a 3D video and formalize the problem as follows. A 3D video is a sequence of $T$ RGB frames $\{I_t\in \mathbb{R}^{3\times H\times W}\}_{t=1}^{T}$ along with their depth maps $\{E_t\in \mathbb{R}^{1\times H\times W}\}_{t=1}^{T}$. Estimating the long-term scene flow amounts to producing 3D trajectories $P^{xyz}_{whole} \in \mathbb{R}^{N \times T \times 3}$ in the camera coordinate system for $N$ query points with known initial positions $P^{xyz}_{query} \in \mathbb{R}^{N \times 1 \times 3}$. 

\textbf{Trajectory initialization} 
The first step of initialization is to divide the entire video into several sliding windows. We partition it with window size $S$ and a sliding step of $S / 2$. As shown on the left side of Fig. \ref{fig: overview}a, we need to track $N$ query points, taking the three green points as examples. 

For the first sliding window, the positions of the trajectory points are initialized as $P^{xyz}_{query}$. For the other sliding windows, the trajectories of first $S / 2$ frames are initialized based on the estimated results of the last $S / 2$ frames of the previous sliding window, and the trajectories of last $S / 2$ frames are initialized based on the estimated result of the last frame of the previous sliding window. 

For any sliding window, we can obtain initialized trajectories $P^{xyz}_{init} \in \mathbb{R}^{N \times S \times 3}$. Furthermore, we transform $P^{xyz}_{init}$ from the camera coordinate system to a $uvd$ coordinate system composed of the $uv$ plane and the depth dimension with camera intrinsics $K=(f_x, f_y, c_x, c_y)$, resulting in initialized trajectories $P^{uvd}_{init} = [f_x \cdot P_{init}^{x} / P_{init}^{z} + c_x, f_y \cdot P_{init}^{y} / P_{init}^{z} + c_y, P_{init}^{z}] \in \mathbb{R}^{N \times S \times 3}$ in the $uvd$ coordinate system. 

\textbf{Downsampling and downscaling} 
The network performs inference at resolution of $h \times w$, where $h = H / s$, $w = W / s$, and $s=8$ is a downsampling factor utilized for efficiency. First, we use a encoder network to extract image features $F = \{F_t\in \mathbb{R}^{c_f\times h\times w}\}_{t=k}^{k+S-1}$ from RGB frames $\{I_t\}_{t=k}^{k+S-1}$ in the current sliding window, assuming that the first frame of the window is the $k$th frame of the entire 3D video. Here, $c_f$ is the number of channels. Encoder is a convolutional neural network comprising 8 residual blocks and 5 downsampling layers. 

Without feature extraction, we directly perform bilinear downsampling with the same factor $s$ on the $S$-frame original depth maps $\{E_t\}_{t=k}^{k+S-1}$, thereby obtaining downsampled depth maps $D = \{D_t\in \mathbb{R}^{1\times h\times w}\}_{t=k}^{k+S-1}$. 
Furthermore, we scale the initialized trajectories $P^{uvd}_{init}$ by a factor of $1/s$ in the $uv$ dimensions to obtain the initial downscaled trajectories $P^{uvd}_{0} = [P_{init}^{u} / s, P_{init}^{v} / s, P_{init}^{d}] \in \mathbb{R}^{N \times S \times 3}$. 

\textbf{Updates of template feature and trajectory} In the flow iteration module (FIM), we iteratively update the template feature and downscaled trajectories of query points. When processing the first frame of the first sliding window, we perform bilinear sampling on image feature $F_1$ using the $uv$ coordinates of the query points to obtain the template feature of $F_1$. We replicate this feature $S$ times along the temporal dimension to obtain the initial template feature $Q_0 \in \mathbb{R}^{N \times S \times c_f}$ for all sliding windows. Each sliding window will have a consistent $Q_0$ and a different $P_0^{uvd}$. 
After $n$ iterations of FIM, they will be updated to $Q_n$ and $P_n^{uvd}$, respectively. 

\textbf{Trajectory output} 
We first scale the updated downscaled trajectories $P_n^{uvd}$ to the original input resolution, producing the 3D trajectory segments $P^{uvd} \in \mathbb{R}^{N \times S \times 3}$ of the current sliding window. 
Then we combine camera intrinsics $K$ to transform $P^{uvd}$ from the $uvd$ coordinate system to the camera coordinate system, resulting in the 3D trajectory segments $P^{xyz} \in \mathbb{R}^{N \times S \times 3}$. 
The conversion process from $P_n^{uvd}$ to $P^{xyz}$ is depicted as a 3D trajectory segment generation module in Fig. \ref{fig: overview}a. 
Finally, we chain the 3D trajectory segments generated by all sliding windows to output the whole 3D trajectories $P^{xyz}_{whole}$. The overlapping part between adjacent windows takes the result from the latter window. 

\subsection{Flow Iteration Module} \label{section: iterator}

Like optical and scene flow, LSFE faces the challenge of large inter-frame displacements. We draw on RAFT \cite{RAFT}, PIPs \cite{pips}, and CoTracker \cite{cotracker}, using an iterative approach to approximate the optimal result, and use Transformers to model long-range dependencies within and between $N$ 3D trajectories of length $S$. Finally, from a 2D perspective, we synchronize the extraction of the appearance correlation feature and the depth residual feature, which unifies the originally fragmented image $uv$ and depth dimensions. Based on the above considerations, we design the flow iteration module to update the template feature and downscaled trajectories, as shown in Fig. \ref{fig: overview}b. 

Taking the $n$th iteration of FIM as an example, the module requires to use $Q_{n-1}$, $F$, $D$, and $P_{n-1}^{uvd}$ to construct the input for a Transformer Updater network, which involves concatenating the appearance correlation feature $C_{n-1}^{uv}$, the depth residual feature $C_{n-1}^{d}$, the appearance motion $O_{n-1}^{uv}$, and the depth motion $O_{n-1}^{d}$. The Transformer Updater network outputs residuals, $\Delta Q_{n}$ and $\Delta P_{n}^{uvd}$, respectively, to update $Q_{n-1}$ and $P_{n-1}^{uvd}$ to $Q_{n}$ and $P_{n}^{uvd}$, serving as the output of the current iteration.

\textbf{Appearance correlation feature} Before constructing the appearance correlation feature $C_{n-1}^{uv}$, an $l$-layer correlation pyramid $Y_{n-1}\in \{\mathbb{R}^{N \times S \times (h/2^{i-1})\times (w/2^{i-1})}\}_{i=1}^l$ between $Q_{n-1}$ and $F$ is calculated by taking the dot product between all pairs of feature vectors at the same frame time and pooling the last two dimensions. 
Then, the network uses $P_{n-1}^{uv}$ to look up $Y_{n-1}$ to obtain $C_{n-1}^{uv}\in \mathbb{R}^{N \times S \times c_a}$. Specifically, $C_{n-1}^{uv}$ is obtained through bilinear sampling of $Y_{n-1}$, layer by layer, with grid size $(2\times r+1)^2$, and concatenating. Here $c_a=l\times (2\times r+1)^2$ and $r$ is the radius of the local neighborhood. 

\textbf{Depth residual feature}
First, we use the downscaled 2D trajectories $P_{n-1}^{uv}$ to perform bilinear sampling on the downsampled depth maps $D$, obtaining the depth sampling trajectories $R_{n-1}^{d} \in \mathbb{R}^{N \times S \times 1}$. Then, we compute inverse depths for $R_{n-1}^{d}$ and the depth prediction trajectories $P_{n-1}^{d}$, and calculate the difference between the two inverse depth trajectories to obtain the depth residual feature $C_{n-1}^{d} \in \mathbb{R}^{N \times S \times 1}$. 

\textbf{Appearance and depth motion} 
We obtain the appearance motion $O_{n-1}^{uv}\in \mathbb{R}^{N \times S \times (2+c_o)}$ and the depth motion $O_{n-1}^{d}\in \mathbb{R}^{N \times S \times 1}$ by calculating the differences in $uv$ positions and depths between each frame and the first frame ($t=k$) of the current window: 
\begin{align}
    O_{n-1}^{uv} &= \left[ P_{n-1}^{uv} - P_{n-1,k}^{uv}, \eta(P_{n-1}^{uv} - P_{n-1,k}^{uv}) \right], \\
    O_{n-1}^{d} &= P_{n-1}^{d} - P_{n-1,k}^{d}.
\end{align}
Here, $\eta(\cdot)$ is the sinusoidal position encoding with $c_o$ channels. 

\subsection{Transformer Updater Network} \label{section: updater}

In the Transformer Updater network, we perform feature enhancements on the concatenated input feature $x_{input} \in \mathbb{R}^{N \times S \times c_t}$ and then predict residuals $\Delta P_{n}^{uvd}$ and $\Delta Q_{n}$, as shown in Fig. \ref{fig: overview}c. Here, $c_t$ is the number of channels. 

First, we construct temporal and spatial sinusoidal position encodings, $\eta_{time}$ and $\eta_{space}$, for $x_{input}$. 
Specifically, $\eta_{time}$ is composed of 1D encodings representing the temporal coordinates, $t\in \{1, ..., S\}$, of all trajectory points, while $\eta_{space}$ comprises 2D encodings representing the reference spatial coordinates of all trajectory points. For the trajectory $\{x_1, ..., x_S\}$, we take the $uv$ coordinates of $x_1$ as the reference spatial coordinates of all points on this trajectory. 
A 1D sinusoidal positional encoding is expressed as
\begin{align}
	\eta_{(coord,2i)} &= sin(coord/10000^{2i/c_t}) \\
	\eta_{(coord,2i+1)} &= cos(coord/10000^{2i/c_t}).
\end{align}
where $coord$ is the coordinate, and $i$ is the channel index. Furthermore, the 2D sinusoidal positional encoding of $uv$ coordinates is obtained by applying 1D encoding separately to each dimension and concatenating the results. 

Then we perform element-wise addition on $x_{input}$, $\eta_{time}$, and $\eta_{space}$ to obtain the encoded input feature $x_{input}^{\eta}$, which then goes through $2\times M$ Transformer Blocks for feature enhancement, with Cross-time and Cross-space Transformer blocks being executed alternately. 
The two types of Transformer blocks divide $x_{input}^{\eta}$ into tokens based on time points and trajectories. After tokenization, both types of Transformer blocks sequentially go through normalization, multi-head self-attention (MHSA), normalization, and multilayer perceptron (MLP) layers, with two skip connections. The model can effectively explore and utilize long-range connections within and between trajectories through the above operations. 

The final Transformer block generates the output feature $x_{output}\in \mathbb{R}^{N \times S \times c_t}$, which is fed into a fully connected layer, whose output can be split into $\Delta P_{n}^{uvd}$ and an intermediate feature $F_{int}\in \mathbb{R}^{N \times S \times c_i}$. $c_i$ is the number of channels. The intermediate feature then goes through normalization, fully connected, and GELU activation layers to ultimately generate $\Delta Q_{n}$. 

\subsection{Training and Inference} \label{section: loss}

Training uses a $T$-frame 3D video as a data unit, with ground truth 3D trajectories of query points to supervise outputs. 

The loss is defined in an unrolled fashion to properly handle semi-overlapping sliding windows. 
Given the ground truth $uv$ positions $P^{uv}_{GT}$ and depths $P^{d}_{GT}$ for any sliding window, we define the loss as follows (for simplicity, we omit the indices of the windows): 
\begin{equation}
    \mathcal{L} = \sum_{\omega=\omega_1}^{\omega_{max}}\sum_{i=1}^{n} \gamma^{n-i} \cdot (||P^{uv}_{i} - P^{uv}_{GT}||_1 + \alpha \cdot ||\frac{1}{P^{d}_{i}} - \frac{1}{P^{d}_{GT}}||_1)
\end{equation}
Here we set $\gamma=0.8$ and $\alpha=250$. $\omega_{1}$ and $\omega_{max}$ are the first and last sliding windows. 

\section{Proposed Dataset} \label{section: dataset}

Generally, sparse annotations are only available for rigid bodies across adjacent frames in real-world SFE datasets \cite{KITTI, waymo}. 
Unlike these, the proposed LSFE dataset requires additional annotations of trajectories that span multiple frames and are not limited to rigid bodies. 
Given a temporal segment of autonomous driving data, we construct a $T$-frame RGB-D video and 3D trajectories of interested points in the first frame. 
Specifically, the interested points are sampled from static backgrounds, moving rigid vehicles, and moving non-rigid pedestrians. 

\subsection{Annotations on Backgrounds}

First, we utilize intrinsic and extrinsic parameters of the camera to extract LiDAR points of the first frame, which can be projected correctly into the image. Then, we use the bounding boxes from 2D object detection to exclude all foreground LiDAR points. Here, we have obtained the first frame's background LiDAR points. 
Taking one LiDAR point $X \in \mathbb{R}^{3}$ as an example, we project it onto the remaining $T-1$ frames using vehicle poses. Formally, the projected point at time $t$ is:
\begin{equation}
    X_{t} = W_{t}^{-1} \cdot W_{1} \cdot X
\end{equation}
Here, $W_{t}$ represents the transformation from the ego vehicle to the world coordinate system at time $t$. 

\subsection{Annotations on Vehicles}

In contrast to the background, the vehicle exhibits its own distinctive motion. To account for this, we follow DriveTrack \cite{drivetrack}, a pioneer in establishing the real-world TAP dataset, to introduce 3D bounding boxes from 3D object tracking, to provide the transformation $B_{t}$ from the world to the box coordinate system at time $t$. 
We use the 3D bounding boxes to filter out all vehicle LiDAR points. Taking one LiDAR point $X \in \mathbb{R}^{3}$ as an example, the projected point at time $t$ is:
\begin{equation}
    X_{t} = W_{t}^{-1} \cdot B_{t}^{-1} \cdot B_{1} \cdot W_{1} \cdot X
\end{equation}

\subsection{Annotations on Pedestrians}

The intricacy and non-rigidity of pedestrian movement determine its annotation difficulties, as evidenced by its absence from current SFE datasets. 
We address this challenge indirectly through the binocular video. 
Firstly, we prepare a $T$-frame segment from a rectified binocular video. 

Then, we employ a semi-automated annotation pipeline to efficiently and accurately label the 2D trajectories of interested points in the left and right view videos. 

\textbf{Labeling the interested point} We develop bespoke annotation software and label the 2D coordinate $x_{1}^{l}$ of the interested point within the left image of the first frame. 

\textbf{Computing the coarse left trajectory} Cotracker \cite{cotracker} is utilized to compute the coarse left trajectory $S^{l}=\{ x_{1}^{l}, ..., x_{T}^{l} \}$ of the interested point in the left-view video. 

\textbf{Computing the coarse right trajectory} LEAStereo \cite{leastereo} is employed to compute disparities $\{ c_{1}, ..., c_{T} \}$ of the interested point, frame by frame. Combining the disparities and $S^{l}$, we derive the coarse right trajectory $S^{r}=\{ x_{1}^{r}, ..., x_{T}^{r} \}$. 

\textbf{Manual refinement phase} The coarse left and right trajectories are displayed in the annotation software, where human annotators will correct any low-quality annotations. 

Finally, we combine the refined left trajectory with the disparity sequence to construct the 3D trajectory. The depth at frame $t$ is:
\begin{equation}
    d_{t} = f_{x} \cdot b / c_{t}
\end{equation}
Here, $b$ is the binocular baseline length. 

\begin{figure*}[t]
	\begin{center}
		\includegraphics[width=\linewidth]{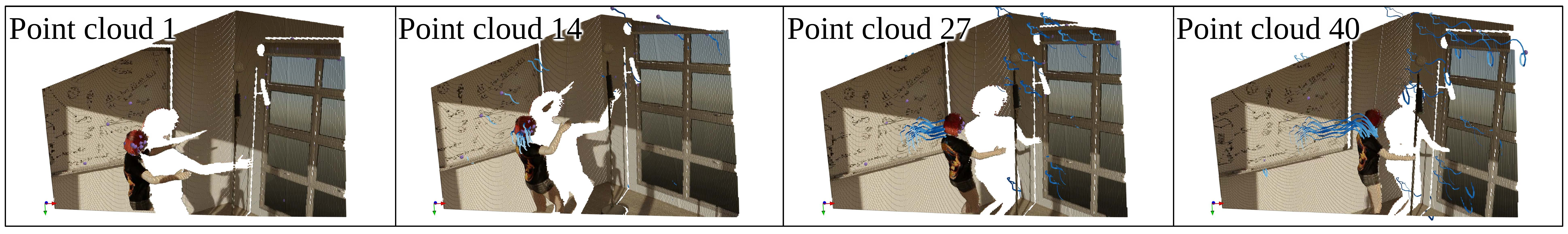}
	\end{center}
	\caption{
		Visualizations of SceneTracker estimation results on the LSFOdyssey test dataset. 
	}
	\label{fig: trajectory}
\end{figure*}

\section{Experiments}

\subsection{Datasets and Experimental Setup}

\subsubsection{Datasets} 

\textbf{LSFOdyssey} We construct an augmented dataset called LSFOdyssey based on PointOdyssey \cite{pointodyssey}, which is a large-scale synthetic dataset and data generation framework for the TAP task. 
The LSFOdyssey dataset comprises 127,437 training data samples acquired through data augmentation techniques such as random flipping, spatial transformations, and color transformations. Each sample includes a 24-frame 3D video along with 3D trajectories of 256 randomly sampled query points. 
We evenly partition 90 testing data samples without applying data augmentation techniques. Each sample consists of 40 frames, including 256 randomly sampled query points. Compared to PointOdyssey, LSFOdyssey additionally performs generation of ground truth 3D trajectories in the camera coordinate system, sequence partitioning, and offline data augmentation. 

\begin{figure}[h]
	\begin{center}
		\includegraphics[width=1.0\linewidth]{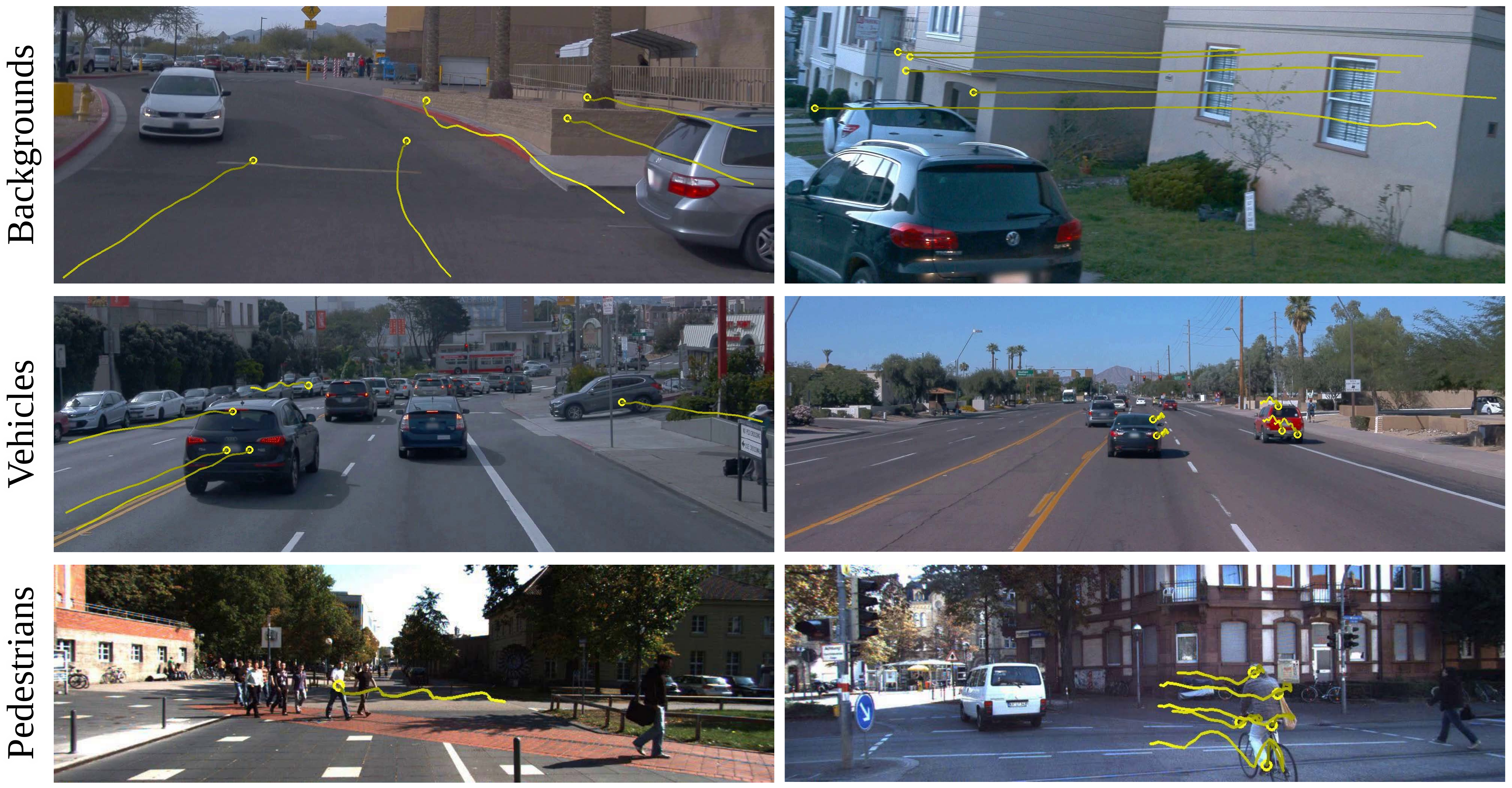}
	\end{center}
	\caption{
		Examples of the proposed LSFDriving dataset. 
	}
	\label{fig: driving}
\end{figure}

\noindent \textbf{LSFDriving} Following the Section \ref{section: dataset}, we construct a real-world evaluation dataset, LSFDriving, for the LSFE field. 
We obtain RGB videos and annotations for backgrounds and vehicles from the Waymo \cite{waymo} dataset. Due to the absence of binocular cameras in Waymo, we apply depth completion methods to generate dense depth sequences. Specifically, we interpolate the nearest neighbors. It is worth noting that we utilize completed depth maps to establish occlusion relationships, thereby providing valid masks for evaluation. 
Meanwhile, we acquire RGB videos for pedestrians from the KITTI \cite{KITTI2012} raw data collection. We utilize LEAStereo \cite{leastereo} to provide dense depth sequences and the proposed semi-automated annotation pipeline to provide annotations. 
We do not offer annotations for left-view occluded and binocular non-co-visible regions. Ultimately, each category comprises 60 testing data samples consisting of 40 frames. There are 5 query points for each background and vehicle sample. 
We annotate 1-5 challenging non-rigid motion points in every pedestrian sample, such as joint and garment points. The examples of LSFDriving are shown in Fig. \ref{fig: driving}

\subsubsection{Evaluation Metrics}

It is common to use metrics the average 2D position accuracy $\delta_{2D}^{avg}$, the median 2D trajectory error $\text{MAE}_{2D}$, and the 2D ``Survival'' rate $\text{Survival}_{2D}$ when evaluating the 2D performance of a predicted 3D trajectory. 
Here, $\delta_{2D}^{avg}$ quantifies the proportion of trajectory points falling within a specified distance threshold from the ground truth, averaged across thresholds $\{1, 2, 4, 8, 16\}$, specified within a normalized resolution of $256\times 256$. 
$\text{Survival}_{2D}^{16}$ is used to evaluate the percentage of frames in which the initial successful tracking occurs; tracking is deemed unsuccessful if the error exceeds 16 pixels in the normalized $256\times 256$ resolution. 

When evaluating the 3D performance of trajectories, we employ the following configuration. 
We utilize $\delta_{3D}^{x}$s and their average value $\delta_{3D}^{avg}$ to measure the percentage of trajectory points within a threshold distance $x \in \{0.10, 0.20, 0.40, 0.80\}$ m to ground truth. We use median 3D trajectory error $\text{MAE}_{3D}$ and 3D end-point error $\text{EPE}_{3D}$ to measure the distance between the estimated and ground truth trajectories. Similarly, we utilize metric $\text{Survival}_{3D}^{0.50}$, with a failure threshold of $0.50$ m.

\subsubsection{Implementation Details}

We use PyTorch to implement our method. We use the AdamW \cite{loshchilov2017decoupled} as an optimizer and the One-cycle \cite{smith2019super} as the learning rate adjustment strategy. 
We set $S=16$, $s=8$, $r=3$, $l=4$, $c_a=196$, and $c_f=c_o=c_i=128$. 
We evaluate our method after 4 FIM iterations, \ie $n=4$, and set $M=6$ for the Transformer Updater network. We use the NVIDIA 3090 GPU for training and inference.

\subsubsection{Training Processes}

Our training includes a training process named Odyssey. 
In the Odyssey training process, we train the model on the LSFOdyssey dataset from scratch. Each data sample learns 24 time frames and 256 query points. The input resolution is $384\times512$. 
The batch size is 8. The learning rate is $2 \times 10^{-4}$, and the number of training steps is 200K. 

\subsection{Visualizations of Results}

Following the Odyssey training process, we visualize the estimation results of SceneTracker on the LSFOdyssey test dataset, as shown in Fig. \ref{fig: trajectory}. 
We present four frames of point clouds from the 40-frame test data sample at equal intervals. 
The predicted 3D trajectories are depicted in blue on their corresponding point clouds. From Fig. \ref{fig: trajectory}, we can see that our method can consistently deliver smooth, continuous, and precise estimation results in the presence of complex motions of the camera and dynamic objects in the scene. 

\begin{figure*}[t]
	\begin{center}
		\includegraphics[width=\linewidth]{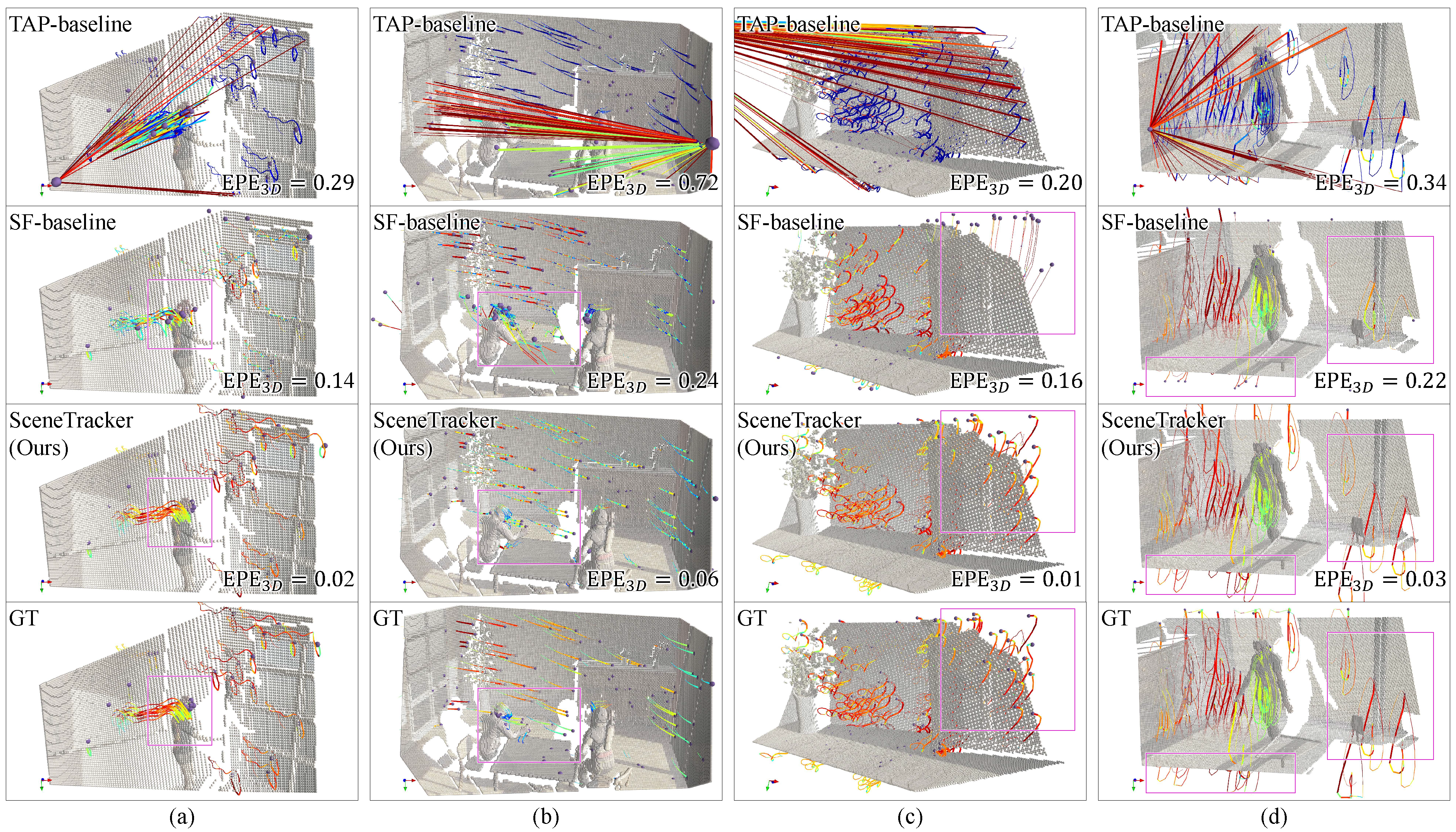}
	\end{center}
	\caption{
		Qualitative results of the TAP baseline, the SF baseline, and our SceneTracker on the LSFOdyssey test dataset. We visualize the trajectory estimates and ground truth of the final frame's point cloud. The trajectories are colorized using a jet colormap. The solid-box marked regions represent areas where the SF baseline exhibits significant errors due to occlusion or exceeding boundaries. 
	}
	\label{fig: odyssey_test}
\end{figure*}

\subsection{Comparisons with the SF baseline}

\begin{table}[h]
	\centering
	\caption{
		Results of 2D metrics on the LSFOdyssey test dataset. 
		All data come from the Odyssey training process. 
	}
	\label{tab: metric_2d}
	\resizebox{0.48\textwidth}{!}{%
		\begin{tabular}{@{}cccc@{}}
			\toprule
			Method & $\delta_{2D}^{avg} \uparrow$ & $\text{Survival}_{2D}^{16} \uparrow$ & $\text{MAE}_{2D} \downarrow$ \\ \midrule
			SF baseline & 72.12 & 91.24 & 5.65 \\
			SceneTracker (Ours) & \textbf{85.07} & \textbf{96.71} & \textbf{2.01} \\ \bottomrule
		\end{tabular}%
	}
\end{table}

\begin{table*}[t]
	\centering
	\caption{
		Results of 3D metrics on the LSFOdyssey test dataset. 
		All data come from the Odyssey training process. 
	}
	\label{tab: metric_3d}
	\resizebox{0.85\textwidth}{!}{%
		\begin{tabular}{@{}ccccccccc@{}}
			\toprule
			Method & $\delta_{3D}^{0.10} \uparrow$ & $\delta_{3D}^{0.20} \uparrow$ & $\delta_{3D}^{0.40} \uparrow$ & $\delta_{3D}^{0.80} \uparrow$ & $\delta_{3D}^{avg} \uparrow$ & $\text{Survival}_{3D}^{0.50} \uparrow$ & $\text{MAE}_{3D} \downarrow$ & $\text{EPE}_{3D} \downarrow$ \\ \midrule
			TAP baseline & 57.52 & 67.72 & 74.45 & 81.24 & 70.23 & 55.76 & 0.388 & 0.573 \\
			SF baseline & 60.66 & 76.35 & 88.39 & 95.11 & 80.13 & 89.92 & 0.176 & 0.188 \\
			SceneTracker (Ours) & \textbf{88.70} & \textbf{95.57} & \textbf{98.08} & \textbf{99.27} & \textbf{95.41} & \textbf{98.09} & \textbf{0.052} & \textbf{0.058} \\ \bottomrule
		\end{tabular}%
	}
\end{table*}

\begin{table*}[t]
	\centering
	\caption{Evaluation results on the LSFDriving dataset, where only points with depth $<35m$ are considered for evaluation. }
	\label{tab: driving}
	\resizebox{\textwidth}{!}{%
		\begin{tabular}{@{}clccccccccc@{}}
			\toprule
			\multirow{2}{*}{Method} & \multicolumn{1}{c}{\multirow{2}{*}{\begin{tabular}[c]{@{}c@{}}Inference\\ Mode\end{tabular}}} & \multicolumn{3}{c}{vehicle} & \multicolumn{3}{c}{background} & \multicolumn{3}{c}{pedestrian} \\ \cmidrule(l){3-11} 
			& \multicolumn{1}{c}{} & $\text{MAE}_{2D} \downarrow$ & $\text{EPE}_{3D} \downarrow$ & $\delta_{3D}^{avg} \uparrow$ & $\text{MAE}_{2D} \downarrow$ & $\text{EPE}_{3D} \downarrow$ & $\delta_{3D}^{avg} \uparrow$ & $\text{MAE}_{2D} \downarrow$ & $\text{EPE}_{3D} \downarrow$ & $\delta_{3D}^{avg} \uparrow$ \\ \midrule
			\multirow{6}{*}{SceneTracker} & ``One'' & 9.13 & 1.595 & 36.45 & 7.00 & 1.548 & 38.36 & 16.38 & 2.902 & 15.48 \\
			& ``One''+loc. & 1.77 & 0.331 & 66.40 & 1.69 & 0.605 & 58.53 & 6.28 & 0.737 & 35.28 \\
			& ``One''+glo. & 1.94 & 0.409 & 55.61 & 1.81 & 0.596 & 58.74 & 5.07 & 0.603 & 39.36 \\
			& ``One''+loc.+glo. & \textbf{1.57} & 0.287 & 65.75 & \textbf{1.45} & 0.472 & 60.88 & 5.49 & 0.707 & 37.98 \\
			& ``All'' & 4.10 & 1.011 & 50.71 & 2.12 & 0.579 & 56.56 & 16.41 & 2.890 & 14.87 \\
			& ``All''+glo. & 1.84 & \textbf{0.252} & \textbf{67.54} & 1.85 & \textbf{0.351} & \textbf{63.87} & \textbf{3.64} & \textbf{0.486} & \textbf{41.32} \\ \midrule
			TAPIR \cite{tapir} & ``One'' & 5.94 & - & - & 5.09 & - & - & 18.18 & - & - \\
			CoTracker \cite{cotracker} & ``All''+glo. & 2.10 & - & - & \textbf{1.00} & - & - & 11.29 & - & - \\
			SpatialTracker \cite{spatialtracker} & ``All''+glo. & \textbf{1.81} & 2.354 & 20.91 & 1.43 & 1.281 & 27.62 & 4.79 & 2.235 & 12.33 \\
			SceneTracker & ``All''+glo. & 1.84 & \textbf{0.252} & \textbf{67.54} & 1.85 & \textbf{0.351} & \textbf{63.87} & \textbf{3.64} & \textbf{0.486} & \textbf{41.32} \\ \bottomrule
		\end{tabular}%
	}
\end{table*}

We use RAFT-3D \cite{raft3d} to design a straightforward scene flow baseline named the SF baseline. 
Since original scene flow methods can only address trajectory prediction problems between two frames, we adopt a chaining approach for long videos. After calculating the scene flow between the first and second frames, we determine the $uv$ coordinates corresponding to the predicted position of the second frame and index the depth of the second frame to obtain the query point coordinates from the second frame to the third frame, and so on for subsequent frames. 

We also train the SF baseline on the LSFOdyssey dataset from scratch to ensure equity. We use the same training settings as the SceneTracker's Odyssey training process. After training, we compare SceneTracker with the SF baseline on the LSFOdyssey test dataset. The results are shown in Fig. \ref{fig: odyssey_test}, TABLE \ref{tab: metric_2d} and TABLE \ref{tab: metric_3d}. 

We can see from TABLE \ref{tab: metric_2d} and TABLE \ref{tab: metric_3d} that our method significantly outperforms the SF baseline across all dataset metrics, both in 2D and 3D metrics. Especially on $\text{MAE}_{2D}$ and $\text{MAE}_{3D}$, SceneTracker achieves error reductions of 64.4\% and 70.5\%, respectively. 
In contrast to the SFE approach, our method employs identical query point features as initial templates tracked for each window, avoiding query point drift, effectively addressing occlusion and out-of-boundary effects, as shown in solid-box marked regions in Fig. \ref{fig: odyssey_test}. 
The experimental findings substantiate the efficacy of our approach in addressing the LSFE challenge.

\subsection{Comparisons with the TAP baseline}

We observe that the 3D trajectories estimated by SceneTracker can be projected onto the image plane to form continuous 2D trajectories, which match the output format of a TAP method. 
Therefore, we construct a TAP baseline based on SceneTracker. 
Specifically, the TAP baseline first outputs the image plane $uv$ trajectory, then utilizes this to index the depth map of each frame, and finally transforms the indexed $uvd$ trajectory into the camera coordinate system. 

We utilize SceneTracker trained through the Odyssey training process to provide 2D trajectories for the TAP baseline. After training, we compare SceneTracker with the TAP baseline on the LSFOdyssey test dataset. The results are shown in Fig. \ref{fig: odyssey_test} and TABLE \ref{tab: metric_3d}. 

We can see from TABLE \ref{tab: metric_3d} that our approach significantly outperforms the TAP baseline on all 3D dataset metrics. Notably, on $\text{MAE}_{3D}$ and $\text{EPE}_{3D}$, SceneTracker achieves error reductions of 86.6\% and 89.9\%, respectively. At the same time, on $\text{Survival}_{3D}^{0.50}$, SceneTracker achieves a 75.9\% increase in accuracy. 
This significant performance gap stems mainly from the lack of robustness of the TAP baseline against depth noise interference and the inability to address occlusion relationships in 3D space. 
It is worth noting that LSFOdyssey simulates fog and smoke, directly setting their depths as outliers, resulting in a depth map with a substantial amount of noise, which poses significant challenges. The TAP baseline, due to its direct indexing from the depth map, is unable to detect and correct erroneous depth data, which is also substantiated in Fig. \ref{fig: odyssey_test}. 

The results fully demonstrate that the proposed method excels in addressing spatial relationships and noise resistance, highlighting the unique challenges posed by the LSFE task. 

\subsection{Results on LSFDriving}

This subsection mainly analyzes the performance of our method on real-world data. The first part of TABLE \ref{tab: driving} shows the evaluation results of SceneTracker under the ``One'' and ``All'' inference modes on the LSFDriving dataset after the Odyssey training process. 
In the ``One'' inference mode, the network tracks only one query point at a time, while in the ``All'' mode, it tracks all query points simultaneously. 
Since each sample in LSFDriving only requires tracking up to five points, even tracking all query points simultaneously (``All'' inference mode) does not provide the network with rich cross-space context. 
Therefore, we enable the network to simultaneously track additional auxiliary points including: (1) local points (+loc.) sampled within the target's 50×50-pixel neighborhood, and (2) global points (+glo.) uniformly distributed across the entire image. 
Following CoTracker \cite{cotracker}, we sample both types of auxiliary points with 6×6 grids. 

It can be seen from TABLE \ref{tab: driving} that adding auxiliary points can significantly enhance network performance, regardless of whether the inference mode is ``One'' or ``All''. 
Simultaneously, the ``All'' inference mode exhibits superior performance across all 3D dataset metrics with auxiliary points. 
Furthermore, we observe that errors in pedestrian samples are consistently higher than those in background and vehicle samples. This showcases the complexity and challenges posed by non-rigid motions. 

Furthermore, the second part of TABLE \ref{tab: driving} compares the performance of our method and related methods including recent TAP methods (TAPIR \cite{tapir} and CoTracker \cite{cotracker}) and a concurrent method SpatialTracker \cite{spatialtracker}, which uses a triplane representation to represent the 3D content of each frame. 
These methods load their respective released weights and use the same ``All''+glo. inference mode (TAPIR only supports the ``One'' inference mode). 
From TABLE \ref{tab: driving}, it can be seen that our method achieves the best results on all 3D metrics and promising results on all 2D metrics. 
It is worth noting that the advantage of SceneTracker
in 3D dataset metrics may be due to the presence of a lot of noise in the depth data of LSFDriving, which interferes with the left and top view projections of SpatialTracker, thereby affecting its final 3D estimation accuracy. 

Overall, our method demonstrates an advantage in generalization abilities in real-world scenes while training solely on synthetic datasets. 

\subsection{Ablation Studies}

\begin{table}[h]
	\centering
	\caption{Results of ablation experiments on the LSFOdyssey test dataset. All data come from the Odyssey training process. }
	\label{tab: ablation}
	\resizebox{0.48\textwidth}{!}{%
		\begin{tabular}{@{}ccccc@{}}
			\toprule
			Experiment & Variation & $\delta_{3D}^{avg} \uparrow$ & $\text{Survival}_{3D}^{0.50} \uparrow$ & $\text{EPE}_{3D} \downarrow$ \\ \midrule
			\multirow{2}{*}{\begin{tabular}[c]{@{}c@{}}Depth residual\\ feature\end{tabular}} & w/o & 91.43 & 97.48 & 0.085 \\
			& w/ & \textbf{95.41} & \textbf{98.09} & \textbf{0.058} \\ \midrule
			\multirow{2}{*}{\begin{tabular}[c]{@{}c@{}}Template\\ feature's updates\end{tabular}} & w/o & 95.27 & 98.04 & 0.061 \\
			& w/ & \textbf{95.41} & \textbf{98.09} & \textbf{0.058} \\ \midrule
			\multirow{3}{*}{\begin{tabular}[c]{@{}c@{}}Window\\ size\end{tabular}} & 8 & 93.32 & 97.77 & 0.075 \\
			& 16 & \textbf{95.41} & \textbf{98.09} & \textbf{0.058} \\
			& 24 & 81.47 & 95.41 & 0.174 \\ \midrule
			\multirow{3}{*}{\begin{tabular}[c]{@{}c@{}}Transformer\\ blocks'\\ number\end{tabular}} & 4 $\times$ 2 & 94.93 & 97.95 & 0.062 \\
			& 6 $\times$ 2 & \textbf{95.41} & \textbf{98.09} & \textbf{0.058} \\
			& 8 $\times$ 2 & 94.10 & 97.82 & 0.067 \\ \bottomrule
		\end{tabular}%
	}
\end{table}

\textbf{Depth residual feature} The depth residual feature calculates an accurate difference between the estimated and measured values of the inverse depth, thus providing an important clue for the estimation of depth residuals. 
We remove the depth residual feature to construct a baseline that uniformly encodes the RGB-D data to establish the connection between the network and the depth input. As shown in Table \ref{tab: ablation}, the addition of the depth residual feature significantly improves the performance of the network on 3D metrics compared with the baseline. 

\textbf{Updates of the template feature} We halt the template feature updates to assess the impact of the dynamic cost volume on network performance. 
The performance degradation induced by the locking of template features in TABLE \ref{tab: ablation} shows that LSFE necessitates a greater focus on addressing the variations in feature distribution brought about by long sequences in comparison to SFE. 

\textbf{Inference sliding window size} TABLE \ref{tab: ablation} shows how the size of the inference sliding window affects model performance. We find that a size of $S=16$ leads to optimal performance across all dataset metrics. 

\textbf{Number of Transformer blocks} We build baselines containing different numbers of Transformer blocks, with $M=4, 6, 8$. As shown in TABLE \ref{tab: ablation}, the baseline with $M=6$ achieves the best performance. 

\begin{figure}[h]
	\begin{center}
		\includegraphics[width=0.9\linewidth]{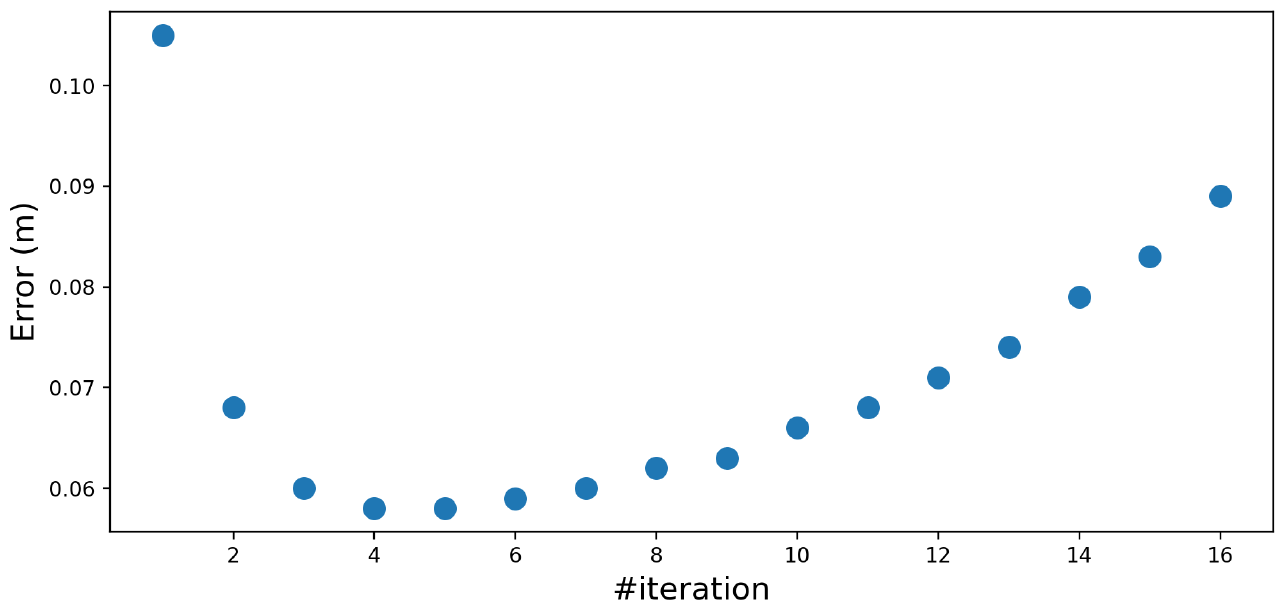}
	\end{center}
	\caption{
		Metric $\text{EPE}_{3D}$ on the LSFOdyssey test dataset versus the number of FIM iterations after the Odyssey training process. 
	}
	\label{fig: update_num}
\end{figure}

\textbf{Number of FIM iterations} 
We evaluate the model with 4 FIM iterations, which gives the best result, as shown in Fig. \ref{fig: update_num}. 

\textbf{Computational efficiency}
When performing inference on a 3D video from the LSFOdyssey test dataset using an NVIDIA 3090 GPU, SceneTracker achieves an inference time of 518 ms, consumes 4.5 GB of VRAM, requires 4163.8G FLOPs of computation, and contains 24.2M parameters. 

\section{Conclusion}
	
In summary, we propose a new task, long-term scene flow estimation, to better capture the fine-grained and long-term 3D motion. 
To effectively address this issue, we introduce a novel approach named SceneTracker. 
With detailed experiments, SceneTracker shows superior capabilities in handling 3D spatial occlusion and depth noise interference, highly tailored to the needs of the LSFE task. Finally, we build a real-world evaluation dataset, LSFDriving, and the experimental results on it further demonstrate SceneTracker's advantage in generalization abilities. 

Through this work, we further prove the feasibility of simultaneously capturing the fine-grained and long-term 3D motion, and the LSFE task is poised to become a promising direction for future research.

\footnotesize
\bibliographystyle{IEEEtran}
\bibliography{IEEEabrv,wangbo}

\end{document}